\documentclass{article}

\PassOptionsToPackage{numbers,compress}{natbib}

\usepackage[dblblindworkshop, final]{neurips_2025}
 \usepackage{adjustbox}
 \usepackage{siunitx}

\usepackage[utf8]{inputenc} 
\usepackage[T1]{fontenc}    
\usepackage{hyperref}       
\usepackage{url}            
\usepackage{booktabs}       
\usepackage{amsfonts}       
\usepackage{amsmath}        
\usepackage{nicefrac}       
\usepackage{microtype}      
\usepackage{xcolor}         
\usepackage{multirow}       
\usepackage{marvosym}
\usepackage{graphicx}
\title{
\vspace{-4em} 
{\fontsize{8}{50}\selectfont{\normalfont{Published at the European Conference on Neural Information Processing Systems (EurIPS 2025), Rethinking AI Workshop}}}\\[3em]
SpikeFit: Towards Optimal Deployment of Spiking Networks on Neuromorphic Hardware
}

\workshoptitle{RethinkingAI}

\author{
    Ivan Kartashov$^{\dagger}$$^{*}$ \quad
    Mariia Pushkareva$^{*}$ \quad
    Iakov Karandashev$^{*}$\\[4pt]
    $^{\dagger}$HSE University\\
    $^{*}$Center of Optical Neural Technologies\\\\
    \small{\textrm{\Letter}}
    \texttt{iakartashov@edu.hse.ru}\\
}

\begin{document}

\maketitle

\begin{abstract}
This paper introduces SpikeFit, a novel training method for Spiking Neural Networks (SNNs) that enables efficient inference on neuromorphic hardware, considering all its stringent requirements: the number of neurons and synapses that can fit on a single device, and lower bit-width representations (e.g., 4-bit, 8-bit). Unlike conventional compressing approaches that address only a subset of these requirements (limited numerical precision and limited number of neurons in the network), SpikeFit treats the allowed weights' discrete values themselves as learnable parameters co-optimized with the model, allowing for optimal Clusterization-Aware Training (CAT) of the model’s weights at low precision (2-, 4-, or 8-bit) which results in higher network compression efficiency, as well as limiting the number of unique synaptic connections to a value required by neuromorphic processor. This joint optimization allows SpikeFit to find a discrete weight set aligned with hardware constraints, enabling the most complete deployment across a broader range of neuromorphic processors than existing methods of SNN compression support. Moreover, SpikeFit introduces a new hardware-friendly Fisher Spike Contrubution (FSC) pruning method showing the state-of-the-art performance. We demonstrate that for spiking neural networks constrained to only four unique synaptic weight values (M = 4), our SpikeFit method not only outperforms state-of-the-art SNNs compression methods and a conventional baselines combining extreme quantization schemes, clustering algorithms, but also meets a wider range of neuromorphic hardware requirements and provides the lowest energy use in experiments. 
\end{abstract}

\section{Introduction}

Spiking Neural Networks (SNNs) are brain-inspired models that promise exceptional energy efficiency when run on specialized neuromorphic hardware \cite{Schuman2022NeuromorphicReview}. Neuromorphic processors have been developed to emulate the event-driven and massively parallel nature of biological neural systems, with prominent examples including Intel’s Loihi (digital asynchronous many-core processor \cite{Davies2018Loihi}), SpiNNaker (ARM-based massively parallel platform with event-routing \cite{Furber2014SpiNNaker}), BrainScaleS (mixed-signal accelerated analog wafer-scale system \cite{Schemmel2010WaferScale}), and mixed-signal chips such as DYNAP-SE2 \cite{dynap}. These platforms promise orders-of-magnitude reductions in energy consumption compared to von Neumann accelerators by co-locating memory and computation and processing information through sparse spikes. Nevertheless, deploying Spiking Neural Networks (SNNs) on current neuromorphic hardware faces stringent architectural and algorithmic constraints. In other words, a significant gap exists between SNNs in simulation and their deployment on physical hardware.

First, neuromorphic hardware offers only limited on-chip memory and connectivity, which restricts the number of neurons and synapses that can be implemented. As a result, networks must often be reduced in size to fit a single device. Second, most platforms support only low-precision or discrete weight values (for example, 2-, 4-, or 8-bit representations), making it difficult to deploy models trained with continuous or high-precision parameters.

Moreover, several neuromorphic systems, designed for density and superior energy efficiency, restrict each synapse to a very small number of available states. For instance, studies on IBM TrueNorth processor \cite{TrueNorth} highlighted its use of just four integer values to represent synaptic weights \cite{Esser2016TrueNorthCNN}. This is a fundamental constraint, reducing a synapse's state to a choice from a small, programmable look-up table. 

Training SNNs to perform well under such extreme constraints is a major challenge. Therefore, further research is required on SNN compression methods that allow full deployment on the hardware.

This paper proposes \textbf{SpikeFit}, an efficient training methodology tailored for producing resource-friendly efficient spiking neural networks deployable on a wide range of neuromorphic hardware. Our method offers another way of reducing the spiking model size, its energy use, and complying with most hardware requirements. SpikeFit method effectively combine efficient, hardware-friendly clusterization, quantization and pruning schemes allowing for a rapid deployment of spiking networks even on the most demanding hardware (e.g. on IBM TrueNorth processor \cite{TrueNorth} requiring to represent SNN's synaptic weights using a fixed set of integer values). The method treats the $M$ allowable discrete weight values not as fixed constants, but as learnable parameters. This can be viewed as a "smart" or adaptive clustering method that operates during training.

The core contributions of this work are:
\begin{enumerate}
    \item[(i)] Clusterization Aware Training (CAT) method finding optimal sets of discrete values to represent weights in low numerical precision where the $M$ unique discrete weight values for each layer are learned end-to-end, allowing the model to discover its own optimal resource-friendly weight representation.
    \item[(ii)] Fisher Spike Contribution (FSC) Pruning method approximating the diagonal Fisher information of channel-wise gates in spiking network.
    \item[(iii)] A comparison focused on models with the same level of precision: our SpikeFit method against baselines: Quantization-Aware Training (QAT) plus clustering and Trained Ternary SNN baselines, versus recent state-of-the-art SNNs compressing approaches.
\end{enumerate}

The key characteristics of SpikeFit are:
a) The values in the codebook (the set of $M$ discrete levels) are not fixed but shift to positions that are most advantageous for minimizing the global loss function
b) The network's underlying latent weights begin to cluster around these learned codebook values as training progresses
c) Optimal clustering technique (CAT) allows to reduce performance loss on post-hoc weight quantization step acting like a quantization regularization technique (making the network less sensetive to noise or drift)
d) The method allows to represent synaptic weights in accordance to available number of unique synaptic states (e.g. 4 available synaptic weight states in IBM TrueNorth \cite{TrueNorth}) d) The Fisher Spike Contribution (FSC) Pruning method precisely removes less important parts of the spiking network according to the Fisher Information of both its spiking activity, and its effect on a global loss function.

\section{Related Work}

Model compression is essential for deploying deep neural networks on resource-constrained devices. Quantization and pruning remain key techniques reducing the precision of model weights, its size, and energy use.

\textbf{Quantization in ANNs} Quantization-Aware Training (QAT) is the standard method for high-performance quantized ANN models, outperforming Post-Training Quantization (PTQ) by simulating quantization in the training loop. Extreme cases include binary and ternary quantization allowing to achieve extreme model's compression via conversion into low numerical precision (1-bit \cite{Wang2023BitNet}, 2-bit \cite{Ma20241BitLLMs, Jain2020BinarizedNN}).

\textbf{Quantization in SNNs} Quantizing SNNs is challenging due to their temporal dynamics and the non-differentiable nature of spiking. While one can convert a trained ANN to an SNN, subsequent quantization can amplify conversion errors. Directly training SNNs using surrogate gradients enables standart QAT-like approaches, although these often focus on standard 4-bit or 8-bit quantization \cite{Lui2021HessianQuantSNN}. Extreme cases like binary \cite{Eshraghian2022DeadNeurons, Kim2020BinarySNN, Shymyrbay2023LowPrecisionSNN} and ternary SNNs use predefined, fixed value sets (e.g., $\{-1, 0, 1\}$ \cite{Sun2022TernarySNN, Amir2017EventGesture}) to achieve extreme compression of the SNN model. The latter methods allow for a fast and complete deployment on a wide range of neuromorphic processors without the use of complicated training schemes, although they often lead to performance degradation \cite{ayan}.

\textbf{Layer-wise Quantization vs Global Quantization}
Layer-wise quantization – i.e. tuning the fixed‐point or reduced‐precision format of each layer separately – reliably outperforms a single global quantization scale because different layers have very different numeric ranges. Moreover, neuromorphic hardware like Intel’s Loihi chips is built for this flexibility – Loihi’s synapses can each use 1–9 bits \cite{Davies2018Loihi} – so quantizing per layer maps directly onto its architecture.

\textbf{Pruning techniques} Pruning in spiking neural networks (SNNs) can be broadly categorized into unstructured and structured approaches. Unstructured pruning removes individual weights or neurons based on local saliency metrics. Magnitude-based strategies \cite{yin} and fine-grained pruning frameworks \cite{shi2024} have demonstrated that extremely sparse connectivity can be achieved without substantial performance degradation. Similarly, biologically inspired schemes such as synaptic rewiring \cite{bellec2017, chen2022} adaptively prune and regrow weak connections. While these methods provide high compression ratios and improved energy efficiency, their irregular sparsity patterns disrupt memory alignment, often hindering hardware-friendliness and requiring specialized hardware for efficient deployment. 

On current neuromorphic processors, unstructured pruning yields little or no hardware benefit because synaptic storage, routing, and device allocation are implemented at coarse-grained levels rather than individual connections. For example, IBM TrueNorth \cite{TrueNorth} implements neurons and synapses in fixed neurosynaptic cores (256 neurons and 64k synapses per core) with event-driven routing tables. Eliminating scattered synaptic weights does not free entries in these tables or reduce reserved per-core synapse banks, making unstructured sparsity ineffective. Similarly, Intel Loihi \cite{Davies2018Loihi} employs hierarchical routing and allocates its 16 MB of on-chip synaptic memory in block formats. Consequently, weight-level sparsification does not reclaim memory or routing resources unless pruning is performed in routing-aware, structured groups.

In contrast, structured pruning eliminates entire channels or kernels, thus preserving the network’s computational regularity and compatibility with off-the-shelf hardware accelerators. The most impactful works on structured pruning combine principled pruning criteria with adaptive or automated strategies, and increasingly consider hardware constraints. The most efficient recent approaches are Spiking Channel Activity-based (SCA) pruning presented in \cite{SCA} and Singular Value of Spike activity pruning criterion (SVS) presented in \cite{Wei2025QPSNN}. However, both methods have issues. SCA-based pruning criterion exhibits a high dependency on inputs. SVS pruning criterion solves the latter problem effectively measuring spatio(-temporal) complexity but ignores the loss signal which can favor redundant patterns or background structures.

\textbf{Compression of SNNs} To further maximize the compression of SNNs recent works include combining pruning and layer-wise quantization. QP-SNN architecture presented in \cite{Wei2025QPSNN} achieves an incredible efficiency by integrating hardware-friendly SVS-based structured pruning technique with quantization technique and weight rescaling strategy applied beforehand. However, this method do not offer a full deployment on neuromorphic processors that limit the number of unique synaptic states (e.g. IBM TrueNorth \cite{TrueNorth} or Kaspersky Alt-AI \cite{Altai}). It is important to also mention its quantization technique (ReScaW) applies rescaling strategy to weights only solving imbalanced weight distribution problem (i.e. long-tailed weight values) out of all weight representation problems affecting uniform quantization accuracy: large dynamic-range differences across layers or channels highlighted in \cite{zhang}, large loss-sensitivity of some groups of weights \cite{dong}. Moreover, weight-only rescaling fix proposed by ReScaW ignore how quantized activations interact with quantized weights (although the joint distribution matters for end-to-end error), as well as ignoring the fact that scalar rounding treats each weight independently, not correlated weight patterns. For comparison, several works addressing similar challenges in LLM weight compression have proposed more effective rescaling strategies. For instance, the Activation-aware Weight Quantization (AWQ) \cite{AWQ} method introduces a simple rescaling scheme based on activation-aware patterns, which appears particularly relevant to our case and yield higher efficiency. In contrast, ReScaW requires manual selection of the gamma parameter based on data-specific insights, which complicates its application and, most importantly, it prevents leveraging activation-aware patterns to achieve state-of-the-art performance. We believe weight representation could be further improved and novel approaches could be proposed to investigate more robust weight representation strategies allowing for better quantization regularization. Specifically, applying clustering algorithms to model's weights has been shown to improve the robustness of ANN models \cite{ClusterRegular}. This makes clustering a promising option for further maximizing SNN model compression, particularly given that clustering methods also align well with certain neuromorphic hardware requirements by limiting available synaptic states. Consequently, there is a clear need to explore new hardware-friendly efficient approaches for SNN compression.

\section{Quantized and Clustered SNN Baseline}
\paragraph{Baseline requirements}
We see the main focus of our contribution is to allow for an optimal and fast SNN compression that is available for a wide range of neuromorphic processors. Therefore, for baseline we take two different classic SNN compression pipelines that have the most complete deployment on the hardware observing requirements: a limited number of neurons in the neural network, a limited set of values (synaptic states) to represent weights, and a lower bit-width representations (e.g., 2-bit). Finally, our baselines are: Ternary Baseline and Clustered Baseline.

\paragraph{The LIF Spiking Neuron Model.}
We employ the widely used Leaky Integrate-and-Fire (LIF) neuron model across all SNN architectures. The continuous-time LIF dynamics describe the membrane potential \(U_{\mathrm{mem}}(t)\) of an RC circuit:
\[
\tau\frac{\mathrm dU_{\mathrm{mem}}(t)}{\mathrm dt}=-U_{\mathrm{mem}}(t)+R\,I_{\mathrm{in}}(t),
\]
where \(\tau=RC\) is the membrane time constant. Using forward Euler discretization with step \(\Delta t\):
\[
U[t+1]=U[t]+\tfrac{\Delta t}{\tau}\!\left(-U[t]+R\,I_{\mathrm{in}}[t]\right).
\]
For deep learning, we set \(\Delta t=R=1\) and absorb parameters into learnable weights \(W\), yielding a discrete form with soft reset:
\[
U[t+1]=\beta U[t]+W X[t+1]-\beta S[t]U_{\mathrm{thr}},
\]
where \(X[t]\) is the presynaptic spike train, \(\beta=e^{-\Delta t/\tau}\) the decay rate, and \(S[t]=\mathbf{1}(U[t]>U_{\mathrm{thr}})\) the output spike \cite{Eshraghian2023SNNTraining}. The soft reset, scaled by \(\beta\), preserves spike timing while enhancing training stability compared to a hard reset.

\paragraph{Implementation of LIF, Simulation, and ANN Conversion Protocols.}
In our experiments, we utilize the LIF neuron with a decay rate of $\beta=0.5$ and a firing threshold of $U_{\mathrm{thr}}{=}1$. To implement SNN versions of ANN architectures ReLU activations are replaced with surrogate‑gradient LIF neurons, pooling and classifiers are applied per time step via SeqToANNContainer from SpikingJelly Python Library \cite{fang}, inputs are replicated across T steps by adding an extra input dimensions, and ResNet shortcuts use time‑distributed projection with pretrained ANN weights remapped into the wrapped modules. To enable backpropagation through the non-differentiable spiking function, we employ an arctangent surrogate gradient.

\paragraph{Clustered Baseline}
One common approach of combining clustering and quantization techniques involves using Post-Training Quantization (PTQ) with weight clustering of a pre-trained, full-precision network using an algorithm like k-Means. However, a stronger baseline, which we use, first fine-tunes a model with Quantization-Aware Training (QAT) and then applies different clustering algorithms.

This baseline approach generates weights for the spiking architecture as follows: (i) train an FP32 CNN, (ii) fine-tune with QAT, (iii) cluster the resulting FP32 layer weights into $M$ centroids (e.g., $k$-means, mini-batch $k$-means, uniform levels) and replace weights by their nearest centroid, (iv) convert to INT8 and briefly fine-tune, and (v) perform ANN$\rightarrow$SNN conversion by dequantizing and loading the clustered/quantized weights into the target SNN. This preserves topology, reduces synaptic states and memory, and yields an SNN readily deployable on neuromorphic hardware.

\paragraph{Ternary Baseline}
Ternary quantization in Spiking Neural Networks assign weights to a discrete set of low-precision values (2-bit) with fixed set of 3 unique values (e.g., $\{-1, 0, 1\}$) observing hardware requirements listed in baseline requirements section. For this reason, we integrated Trained Ternary Quantization algorithm presented in \cite{TTQuant} for our SNN architectures for further method evaluation.

\section{SpikeFit Method}
When deploying SNNs on recent neuromorphic devices researchers face many limitations on neural network size, its weights representation, energy use. This makes SNNs compressing techniques such as quantization and pruning a promising approach to comply with the limitations. Existing research describes different pruning and quantization schemes to comply with: a limited number of neurons in the neural network, a lower bit-width representations (e.g., 4-bit, 8-bit), limitations on energy use. However, listed limitations are a small subset of all requirements that neuromorphic systems list. For instance, the core architecture of the widely known neuromorphic chip IBM TrueNorth lays out in binary synapse connectivity and 4 axon types. Thus, it supports 4 various synaptic weight values and requires proposing a LUT of 4 unique values per neuron \cite{lutnorth}. Some other neuromorphic platforms also report a limitation of a fixed set of synaptic states: Kaspersky Alt-AI \cite{Altai} (limits synaptic weight to 4 unique values), Darwin-3 \cite{darwin} (use axon or weight indexes to encode topology and reduce the per-synapse memory footprint). Hybrid processors report similar requirements (e.g. Tianjic chip \cite{tianjic} uses a weight-index (codebook) approach in its FCore). Moreover, biologically plausible spiking networks (mostly used in tasks related to neuroscience field, computational simulations) have shown to implement synapses with a finite number of discrete weight states (e.g. 2-, 4-, 8- or ~16-state synapses) that are sufficient (and sometimes necessary) for memory, learning, dynamics, and plasticity to behave in biologically realistic ways, while reducing weight‐storage and compute cost \cite{barrett2008, woods2015, dasbach2021} which is a very similar limitation that will require proposing a set of discrete values if the network weights were not directly trained. Therefore, all the latter cases require special techniques to comply with the restrictions. SpikeFit offers a novel method combining state-of-the-art quantization, pruning and clusterization schemes directly for spiking networks training to run inference on a wide range of processors achieving state-of-the-art performance and maintaining resource-friendly efficiency.
\subsection{Clusterization Aware Training (CAT)}

Inspired by Vector Quantization (VQ) techniques, popularized by the Vector-Quantized Variational Autoencoders (VQ-VAEs) \cite{VQVAE} we propose Clusterization Aware Training (CAT) method for efficient compression of the spiking model, reducing weight numerical precision and overcoming the neuromorphic hardware requirement of using a fixed set of values to represent weights. CAT is a clusterization-aware training technique that learns the optimal set of $M$ discrete values for the network weights, rather than imposing a fixed structure like most clusterization algorithms. This allows the network to adapt its weight representation to the specific task and architecture, build its own regularization scheme that yields the robust weight representation for post-hoc quantization along the way.

\paragraph{Learnable Codebooks and Latent Weights}
For each layer that uses Clusterization Aware Training (CAT), we maintain two sets of parameters:
\begin{enumerate}
    \item \textbf{Latent Weights ($W_{\text{latent}}$):} A full-precision weight tensor, identical to weights in a standard neural network. These are updated via backpropagation and represent the "ideal" weights the network aims to learn.
    \item \textbf{Codebook ($C$):} A small, learnable vector $C = (c_1, c_2, \dots, c_M)$ of $M$ floating-point values. This vector, unique to each layer, defines the entire set of discrete values that the final quantized weights can take. The codebook itself is a learnable parameter.
\end{enumerate}

\paragraph{Weight Transformations and Gradient Flow}
The core idea is to map each latent weight to the closest value in the codebook during the forward pass, while ensuring gradients can flow back to both the latent weights and the codebook during the backward pass.

\begin{figure}[h!]
    \centering
    \makebox[\textwidth][c]{%
        \includegraphics[width=1.1\linewidth]{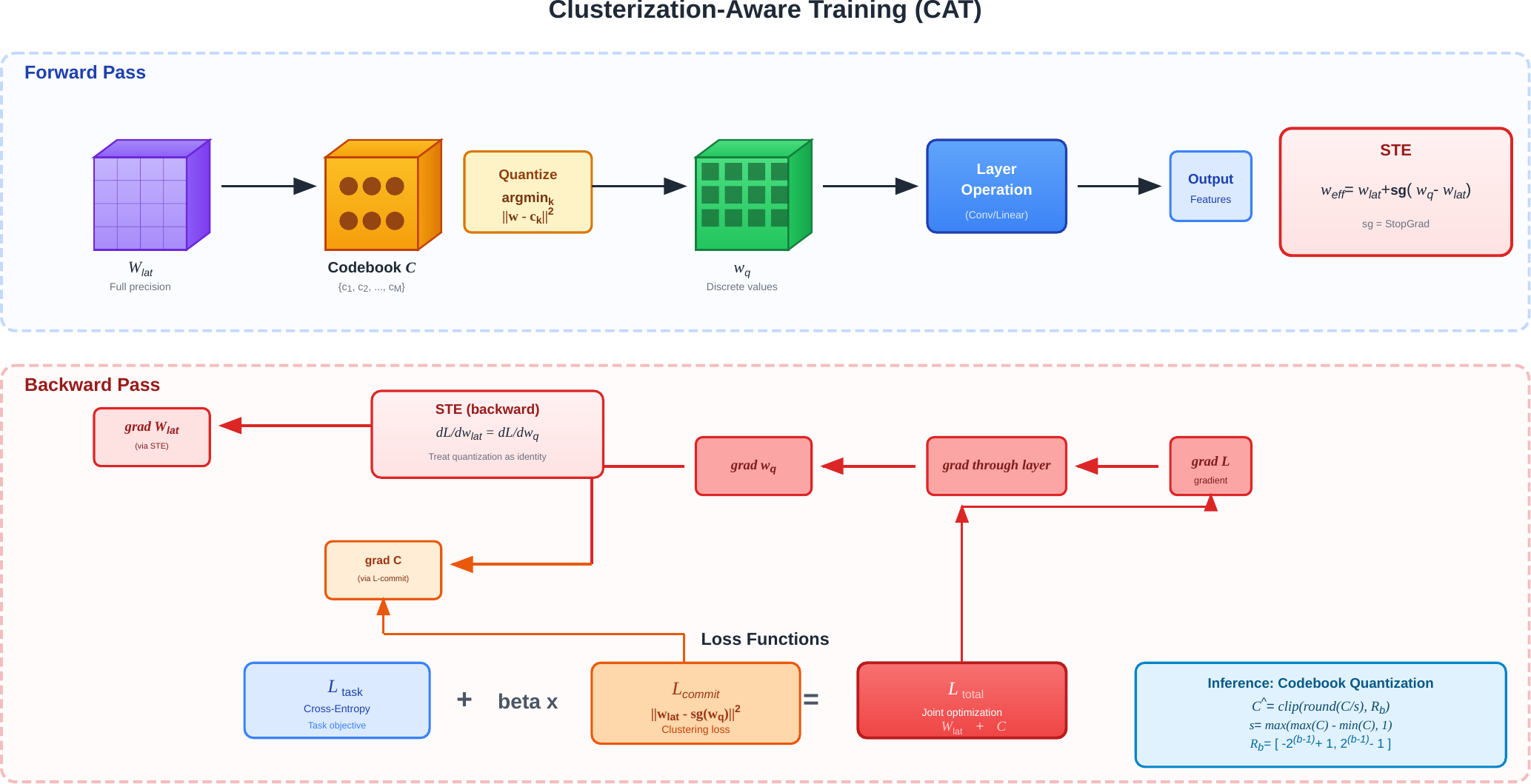}
    }
    \caption{Overview of the Clusterization Aware Training (CAT) pipeline.}
    \label{fig:cat_pipeline_viz_safe}
\end{figure}

\textbf{Forward Pass:} Each latent weight $w^{\text{latent}}$ is quantized to its nearest neighbor in the codebook $C$. The resulting weight, $w^{\text{quantized}}$, is used for the layer's operation (e.g., convolution).
\begin{equation}
    k^* = \arg\min_{k \in \{1,\dots,M\}} \|w^{\text{latent}} - c_k\|^2
\end{equation}
\begin{equation}
    w^{\text{quantized}} = c_{k^*}
\end{equation}

\textbf{Backward Pass:} The $\arg\min$ operation is non-differentiable. To overcome this, we use the Straight-Through Estimator (STE) \cite{Bengio2013STE}. For the backward pass, the gradient is passed directly to the latent weights $W_{\text{latent}}$ as if no quantization occurred. This is achieved with the following formulation for the effective weight $W_{\text{eff}}$ used in computation:
\begin{equation}\label{eq:ste_Clusterization Aware Training (QAT)_main_paper}
    w^{\text{eff}} = w^{\text{latent}} + \text{StopGradient}(w^{\text{quantized}} - w^{\text{latent}}),
\end{equation}
where StopGradient indicates that gradients are not propagated through the the term when computing this loss.
This trick ensures that gradients with respect to the loss flow to $W_{\text{latent}}$. The codebook values $c_k$ are part of the computation graph and are updated directly.

\paragraph{Composite Loss for Smart Clustering}
Training is guided by a composite loss function. The first component is the standard task loss, $\mathcal{L}_{\text{task}}$ (e.g., Cross-Entropy). The second component is a \textbf{commitment loss}, $\mathcal{L}_{\text{commit}}$, which enables the "smart clustering" behavior. It penalizes the distance between the latent weights and their corresponding chosen codebook values:
\begin{equation}\label{eq:commit_loss_Clusterization Aware Training (QAT)_main_paper}
    \mathcal{L}_{\text{commit}} = \frac{1}{N_w} \Big\| w^{\text{quantized}} - \text{StopGradient}(w^{\text{latent}}) \Big \|^2_2,
\end{equation}
where $N_w$ is the number of weights in the layer (e.g., a convolutional layer), and StopGradient indicates that gradients are not propagated through $w^{\text{latent}}$ when computing this loss.
This loss encourages each codebook entry $c_k$ to move toward the centroid of the latent weights assigned to it. By stopping the gradient on $w^{\text{latent}}$, only the codebook $C$ is updated during backpropagation. As a result, the codebook vectors $c_k$ adapt to represent optimal cluster centers, and the latent weights $W_{\text{latent}}$ are indirectly guided to cluster around them.

The final loss function is a weighted sum:
\begin{equation}\label{eq:total_loss_Clusterization Aware Training (QAT)_main_paper}
    \mathcal{L}_{\text{total}} = \mathcal{L}_{\text{task}} + \beta \cdot \mathcal{L}_{\text{commit}}
\end{equation}
where $\beta$ is a hyperparameter balancing the two objectives. By minimizing $\mathcal{L}_{\text{total}}$, we jointly optimize the latent weights and the codebooks.

\paragraph{Codebook Quantization}  
To reduce memory footprint and accelerate inference, we quantize the codebook into a low-bit integer representation.  

For a codebook $\mathbf{C}$, we first compute the dynamic range and corresponding scaling factor:
\[
s = \max \bigl( \max(\mathbf{C}) - \min(\mathbf{C}),\, 1 \bigr).
\]

Given a target bit-width $b$, the representable integer range is
\[
\mathcal{R}_b = \bigl[ -2^{b-1}+1, \, 2^{b-1}-1 \bigr].
\]

The quantized codebook is then obtained as
\[
\hat{\mathbf{C}} = \mathrm{clip}\!\left( \mathrm{round}\!\left(\tfrac{\mathbf{C}}{s}\right),\, \min(\mathcal{R}_b),\, \max(\mathcal{R}_b) \right).
\]

During inference, the quantized weights are reconstructed as $w^{\text{quantized}} = s{\hat{c}}_{k^*}$ where $k^* = \arg\min_{k \in \{1,\dots,M\}} \|w^{\text{latent}} - s\hat{c}_k\|^2$.
    
\subsection{Fisher Spike Contribution (FSC) Pruning}
\label{sec:fsc}

\paragraph{Problem analysis.}
Pruning in spiking neural networks (SNNs) can be broadly categorized into unstructured and structured approaches. Unstructured pruning removes individual weights or neurons based on local saliency metrics. Magnitude-based strategies \cite{yin} and fine-grained pruning frameworks \cite{shi2024} have demonstrated that extremely sparse connectivity can be achieved without substantial performance degradation. Similarly, biologically inspired schemes such as synaptic rewiring \cite{bellec2017, chen2022} adaptively prune and regrow weak connections. While these methods provide high compression ratios and improved energy efficiency, their irregular sparsity patterns disrupt memory alignment, often requiring specialized hardware for efficient deployment.
Structured pruning has long leveraged curvature-aware criteria to minimize the loss increase after parameter removal.
Classical methods such as Optimal Brain Damage (OBD) \citep{lecun1990obd} and Optimal Brain Surgeon (OBS) \citep{obs} use (diagonal or exact) second-order information to estimate deletion of certain groups of parameters' impact. Similarly, first-order Taylor pruning \citep{molchanov2017pruning} captures loss sensitivity, although might be brittle due to sign cancellations.

For spiking networks, surrogate-gradient learning \citep{neftci2019surrogate} enables backpropagation through non-differentiable spike functions, which makes gradient-based saliencies feasible.
Therefore, we came up with a novel pruning criterion motivated by desire to include computationally cheap loss alignment into the pruning scheme reinforced by the equivalence between the expected Hessian and the Fisher information under regularity conditions \citep{amari1998natural, wood, martens}, and by the wish to exploit temporal coding in SNNs that static activity measures (e.g., Spiking Channel Activity (SCA) criterion \citep{SCA}, Singular-Value Saliency (SVS) criterion \citep{Wei2025QPSNN}) may overlook by averaging pruning criterion over time steps \citep{Wei2025QPSNN}.
Recent pruning works exploring gradient flow at initialization (e.g., SynFlow \citep{tanaka2020synflow}) further support using label-aware sensitivity. Our proposed FSC criterion is a label-aware, curvature-aligned pruning criterion adapted to SNNs. FSC retains channels with high loss sensitivity integrated over time, which better preserves accuracy at a fixed pruning ratio.

\paragraph{Problem setup.}
Consider a spiking layer with output activations $y \in \mathbb{R}^{N \times T \times C \times H \times W}$ (batch $N$, time $T$, channels $C$, spatial $H{\times}W$).
Let $L$ denote the task loss computed on logits downstream of this layer.
Define the backpropagated gradient tensor $\delta := \partial L / \partial y$ with the same shape as $y$.
We write $\delta_{b,t,c,h,w}$ and $y_{b,t,c,h,w}$ for sample $b$, time $t$, channel $c$ and spatial location $(h,w)$.

\begin{figure}[htbp]
    \centering
    \makebox[\textwidth][c]{%
        \includegraphics[width=1.1\linewidth]{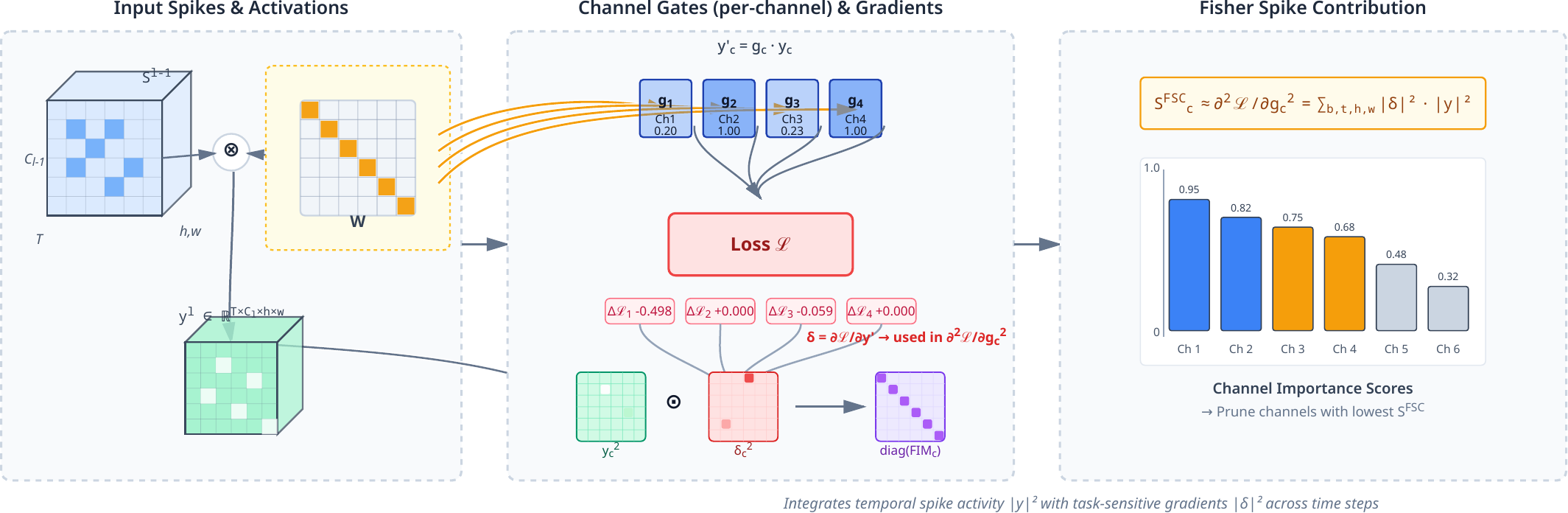}
    }
    \caption{Overview of the Fisher Spike Contribution (FSC) pruning pipeline.}
    \label{fig:fsc}
\end{figure}

\paragraph{Channel-gate view and Fisher connection.}
Following the classic pipeline for estimation of parameters contribution to the global loss function in response to parameters deletion expressed by objective function incrementation mentioned in \cite{lecun1990obd}, let's introduce a multiplicative channel gate $g_c$ acting on the output channel $c$:
\[
y'_{b,t,c,h,w} \;=\; g_c \, y_{b,t,c,h,w}.
\]
The objective function (the global loss function) could be approximated by a Taylor series. We would like to estimate the curvature property change in response to parameters deletion. Thus, for a small perturbation of $g_c$ around $1$, we estimate a second-order expansion of the loss the following way:
\[
\Delta L \;\approx\; (g_c - 1) \,\frac{\partial L}{\partial g_c}
\;+\; \frac{1}{2} (g_c - 1)^2 \,\frac{\partial^2 L}{\partial g_c^2}.
\]
By the chain rule,
\[
\frac{\partial L}{\partial g_c}
\;=\; \sum_{b,t,h,w} \delta_{b,t,c,h,w}\, y_{b,t,c,h,w}.
\]
Since computing and storing the Hessian
matrices directly is prohibitive, we will focus on efficient ways to approximate it.
Under standard conditions for negative log-likelihood losses, the expected Hessian equals the Fisher information matrix \citep{amari1998natural}.
In order to prune individual channels,  it is necessary and sufficient to have a scalar importance per parameter. The (diagonal) Fisher for gate $g_c$ can be approximated as
\[
F_{cc} \;\approx\; \mathbb{E}\!\left[\left(\frac{\partial L}{\partial g_c}\right)^2\right]
\;=\; \mathbb{E}\!\Bigg[\Big(\sum_{b,t,h,w} \delta_{b,t,c,h,w}\, y_{b,t,c,h,w}\Big)^2\Bigg].
\]
Expanding the square and neglecting cross-terms (or upper-bounding by Cauchy–Schwarz), a computable and robust surrogate is
\[
F_{cc} \;\approx\; \kappa \,\mathbb{E}\!\Big[\sum_{b,t,h,w} \delta_{b,t,c,h,w}^{\,2}\, y_{b,t,c,h,w}^{\,2}\Big],
\]
for some constant $\kappa$ depending on the number of terms and correlations.
This leads directly to our FSC saliency.

\paragraph{FSC saliency.}
We define the Fisher Spike Contribution score for channel $c$ as
\begin{equation}
\label{eq:fsc}
S_c^{\text{FSC}}
\;=\; \frac{1}{N} \sum_{b=1}^{N} \sum_{t=1}^{T} \sum_{h=1}^{H} \sum_{w=1}^{W}
\big(\delta_{b,t,c,h,w}\big)^{2}\, \big(y_{b,t,c,h,w}\big)^{2}.
\end{equation}
Intuitively, $S_c^{\text{FSC}}$ is large only when a channel spikes (large $|y|$) exactly where/when the loss is sensitive (large $|\delta|$).

\begin{table*}[h]
\centering
\caption{Comparison of pruning and compression methods across networks and strategies.
Bold blue indicates our proposed FSC method.}
\label{tab:pruning_results}
\resizebox{\textwidth}{!}{%
\begin{tabular}{l l l c c c c c c}
\toprule
\textbf{Method} & \textbf{Network} & \textbf{Pruning} &
\textbf{Size (MB)} & \textbf{Energy (mJ)} &
\textbf{Acc (\%)} & \textbf{Prec (\%)} & \textbf{Rec (\%)} & \textbf{F1 (\%)} \\
\midrule
\multirow{3}{*}{QP-SNN} & \multirow{3}{*}{VGG-16} & SVS \cite{Wei2025QPSNN}         & 14.723 & \textbf{0.03} & 86.656$\pm$0.116 & 86.842$\pm$0.107 & 86.656$\pm$0.116 & 86.628$\pm$0.118 \\
 & & SCA \cite{SCA}           & 14.723 & \textbf{0.03} & 83.275$\pm$0.37 & 84.927$\pm$0.27 & 83.278$\pm$0.37 & 83.373$\pm$0.39 \\
 & & \textbf{\textcolor{blue}{FSC (ours)}} & 14.725 & \textbf{0.03} & \textbf{87.546$\pm$0.059} & \textbf{87.621$\pm$0.073} & \textbf{87.546$\pm$0.059} & \textbf{87.553$\pm$0.065} \\
\bottomrule
\end{tabular}%
}
\end{table*}

\paragraph{Why FSC performs best in our experiments.}
Empirically, FSC consistently outperforms activity- or rank-based criteria (e.g., SVS) across sparsity regimes in our QP-SNN setting.
We hypothesize the reason is the
\textbf{loss alignment}: FSC is directly proportional to (a diagonal approximation of) the Fisher information $F_{cc}$ of a channel gate, hence it targets channels whose removal minimally increases expected loss, unlike SVS which is label-agnostic and rather measures representational complexity, but not task relevance. Two channels with similar SVS score can have vastly different impacts on the loss: one may spike during decisive moments (large $|\delta|$), while another spikes in background regions (small $|\delta|$).

\section{Experiment}

\paragraph{SNN Deployment Evaluation}
There are many efficient SNN compression methods introduced by other authors allowing to successfully deploy spiking models on specialized hardware. However, many of the works lack clear and objective evaluation methods as most experiments include task performance metrics only (accuracy, F1, and similar) showing whether the model handles the task well, but not representing the degree of improvement of the SNN model’s performance on neuromorphic hardware. To estimate a success of both compression and deployment of SNN models we propose a novel metric and include it in our experiments.

\paragraph{DeployRatio: A Composite Deployability Metric}
We propose \textbf{DeployRatio (DR)}, a simple composite metric to quantify the deployability of Spiking Neural Networks (SNNs) by jointly capturing task performance, latency, energy, and model footprint:
\[
\mathrm{DR}_x=\frac{\mathrm{Perf}_x}{L\,E\,S}, \quad x\in\{\text{acc},\text{f1}\},
\quad
\mathrm{Perf}_{\text{acc}}=\tfrac{\text{Accuracy}[\%]}{100},\;
\mathrm{Perf}_{\text{f1}}=\tfrac{\text{F1}[\%]}{100}.
\]
Here \(L\) is inference latency (s/input), \(E\) energy per inference (mJ), and \(S\) model size (MB). A higher DR indicates a model that is accurate, fast, energy-efficient, and compact. The components are:
\begin{itemize}
\item \textbf{Task performance:} \(\mathrm{Perf}_x\in[0,1]\) from accuracy or macro-F1 (\%/100).
\item \textbf{Latency:} \(L=\frac{1}{B}\sum_{b=1}^{B}\Delta t_b\), mean inference time over \(B\) batches.
\item \textbf{Model size:} \(S=\frac{P\,b}{8\!\times\!10^{6}}\) MB, with parameter count \(P\) and precision \(b\) (bits/weight).
\item \textbf{Energy:} \(E\) (mJ) estimated from operation counts and per-operation energy constants.
\end{itemize}
DR is dimensionless and monotonic—higher with better performance and lower with increased latency, energy, or size. \(\mathrm{DR}_{\text{acc}}\) and \(\mathrm{DR}_{\text{f1}}\) target different objectives (balanced vs. imbalanced classes).

\begin{table*}[h]
\centering
\sisetup{
  separate-uncertainty = true,
  table-format = 2.2(2)
}
\setlength{\tabcolsep}{6pt}
\renewcommand{\arraystretch}{1.25}
\caption{
  Performance comparison of compression methods. Accuracy is reported as \textbf{mean ± std} over five runs.
  \textbf{DR (DeployRatio)} indicates the deployment cost-performance ratio.
  Bold blue indicates our proposed \textbf{SpikeFit} results.
}
\label{tab:results}
\begin{adjustbox}{width=\textwidth,center}
\begin{tabular}{
  l l l
  S[table-format=1.0]
  S[table-format=2.2]
  S[table-format=3.2]
  S[table-format=1.0]
  S[table-format=2.2]
  S[table-format=2.2]
  l
}
\toprule
\textbf{Dataset} & \textbf{Network} & \textbf{Method} & {\textbf{Precision}} & {\textbf{Size (MB)}} & {\textbf{Energy (mJ)}} & {\textbf{Timestep}} & {\textbf{Accuracy (\%)}} & {\textbf{F1 (\%)}} & {\textbf{DR (\%)}} \\
\midrule
\multirow{4}{*}{CIFAR-10} & \multirow{4}{*}{VGG-16} & Clustered Baseline & 8 & 14.723 & \textbf{0.004} & 4 & {77.57$\pm$0.49} & {77.54$\pm$0.44} & 68.649 \\
 & & Ternary Baseline & 8 & 14.723 & {0.014} & 4 & {76.244$\pm$0.12} & {76.51$\pm$0.04} & 22.051 \\
 & & QP-SNN & 8 & 14.723 & {0.032} & 4 & {86.58$\pm$0.32} & {86.91$\pm$0.32} & 10.883 \\
 & & \textbf{\textcolor{blue}{SpikeFit (ours)}} & 8 & 14.725 & \textbf{0.004} & 4 & \textbf{89.14$\pm$0.82} & \textbf{89.99$\pm$0.84} & \textbf{89.640} \\
\bottomrule
\end{tabular}
\end{adjustbox}
\end{table*}

\subsection{Analysis of Results}
The evaluation results across all experiments (Table \ref{tab:results}) prove a consistent result on SpikeFit-optimized architecture (VGG-16) performing best and showing state-of-the-art performance in the widely-known benchmark of multi-label classification on CIFAR-10 dataset. Moreover, SpikeFit allows for the most complete deployment of the SNN models by considering the limitation on unique synaptic states number required by some neuromorphic processors that focus on extreme energy efficiency (e.g. IBM TrueNorth) and showing the best performance according to the proposed DeployRatio (DR) metric.

\section{Conclusion}
SpikeFit training methodology demonstrated that combining a novel Clusterization-Aware-Training (CAT) method with quantization and the Fischer Spike Contribution (FSC) structured pruning scheme might be an effective way to deploy hardware-friendly SNNs that operate fast, with low-energy use at state-of-the-art performance. Hopefully, our contribution will motivate researchers to introduce methods offering even better energy efficiency and performance on a wide range of neuromoprhic processors.

\section{Ablation Study}
To further prove the effectiveness of SpikeFit training methodology, we conduct extensive ablation studies. We perform thorough ablation experiments to validate the effectiveness of the proposed CAT weight clusterization strategy and FSC-based pruning criterion. All ablation experiments are conducted on the CIFAR-10 dataset using VGG-16 architecture.

We compare the performance of quantized SNNs (not involve pruning process) with different M settings to evaluate performance on a really small codebook. As depicted in Table \ref{tab:ablation}, the quantized (clustered) SNN using M=2 (8-bit weight clustered to 2 unique values) outperforms state-of-the-art QP-SNN method combining ReScaW quantization strategy with SVS pruning scheme storing weights in 8-bit precision. After increasing the size of the codebook of CAT by setting M=4 the performance is significantly improved by 2.44\% providing even better results showing superior performance and surpassing ReScaW quantization scheme results (without pruning). This proves CAT effectivness when compared to state-of-the-art ReScaW quantization method. \cite{Wei2025QPSNN}

\begin{table*}[h]
\centering
\sisetup{
  separate-uncertainty = true,
  table-format = 2.2(2)
}
\setlength{\tabcolsep}{6pt}
\renewcommand{\arraystretch}{1.25}
\caption{
  Ablation study on VGG-16 SNN architecture on the effectiveness of proposed CAT method.
}
\label{tab:ablation}
\begin{adjustbox}{width=\textwidth,center}
\begin{tabular}{
  l l l
  S[table-format=1.0]
  S[table-format=2.2]
  S[table-format=1.0]
  S[table-format=2.2]
  S[table-format=2.2]
  l
}
\toprule
\textbf{Dataset} & \textbf{Network} & \textbf{Method} & {\textbf{Precision}} & {\textbf{Size (MB)}} & {\textbf{Timestep}} & {\textbf{Accuracy (\%)}} & {\textbf{F1 (\%)}} \\
\midrule
\multirow{5}{*}{CIFAR-10} & \multirow{5}{*}{VGG-16} & Full-precision & 32 & 58.91 & 4 & \textbf{90.97} & \textbf{90.92} \\
 & & ReScaW \cite{Wei2025QPSNN}& 8 & 14.72 & 4 & 89.14 & 89.73\\
 & & QP-SNN \cite{Wei2025QPSNN} & 8 & 14.72 & 4 & 86.94 & 86.92 \\
 & & CAT (M=2) & 8 & 14.72 & 4 & 87.56 & 85.24 \\
 & & CAT (M=4) & 8 & 14.72 & 4 & \textbf{90.00} & \textbf{89.99} \\
\bottomrule
\end{tabular}
\end{adjustbox}
\end{table*}

\section{Experiment Setup}
\subsection{Training and Optimization Details}\label{s:exp_details_main}
We summarize the training hyperparameters for each dataset in Table \ref{tab:results}, including time step, image resolution, optimizer, and other factors.
\begin{itemize}
    \item \textbf{Optimizers and Schedulers:} AdamW with CosineAnnealingLR for all stages; weight decay $1\!\times\!10^{-5}$; $\eta_{\min}=0$; $T_{\max}$ equals the stage's epoch count.
    \item \textbf{Loss Functions:} Cross-entropy for task loss. For SpikeFit, an additional codebook commitment loss is used, scaled by $\beta_{\text{commit}}=0.5$.
        \item \textbf{Pipeline schedule:}
    \begin{itemize}
        \item FP32 stage: $300$ epochs, initial LR $1\!\times\!10^{-2}$.
        \item ReScaW (SNN) stage: $150$ epochs fine-tuning, initial LR $1\!\times\!10^{-4}$, $T{=}4$.
        \item CAT stage: $150$ epochs fine-tuning, initial LR $1\!\times\!10^{-4}$, $T{=}4$, $M{=}4$, codebook quantization=True, bias=False.
    \end{itemize}
    \item \textbf{SNN Parameters:} LIF neuron with $\tau{=}0.5$ (a.k.a., $\beta$) and threshold $1.0$; $T{=}4$ timesteps.
    \item \textbf{Number of Unique Values ($M$):} $M{=}4$ for CAT codebooks.
    \item \textbf{Dataset \& Resolution:} CIFAR-10 at $32{\times}32$.
    \item \textbf{Batch Size \& Workers:} Batch size $1256$; $8$ data-loader workers.
    \item \textbf{Data Augmentation (CIFAR-10):} RandomCrop$(32,\text{padding}{=}4)$, RandomHorizontalFlip, CIFAR10Policy (AutoAugment); normalization mean $(0.4914, 0.4822, 0.4465)$, std $(0.2023, 0.1994, 0.2010)$.
    \item \textbf{Pruning (QP-SNN):} prune ratio $0.3$; SVS limit $5$; SVD rank $k{=}3$; latency/energy trade-offs $\lambda_{\text{lat}}{=}0.2$, $\beta_{\text{energy}}{=}0.1$;
    \item \textbf{Evaluation Protocol:} Each configuration evaluated over $5$ validation repeats.
    \item \textbf{Deployment/Proxyf Metrics (DR settings):} Hardware energy mapping auto-selected; MAC energy $4.6$ pJ, accumulate energy $0.9$ pJ; spike rate $0.2$; latency estimated over $3$ batches.
\end{itemize}

\end{document}